\documentclass[10pt]{article}

\usepackage{iftex}
\ifPDFTeX
  \usepackage[utf8]{inputenc}
  \usepackage[T1]{fontenc}
  \usepackage{lmodern}
\else
  \usepackage{fontspec}
\fi

\usepackage[nopatch=footnote]{microtype}
\usepackage[top=2.5cm,bottom=2.8cm,left=2.5cm,right=2.5cm]{geometry}
\usepackage{amsmath,amssymb}
\usepackage{graphicx}
\usepackage{booktabs}
\usepackage{array}
\usepackage{tabularx}
\usepackage{longtable}
\usepackage{multirow}
\usepackage[table]{xcolor}
\usepackage{enumitem}
\usepackage{caption}
\usepackage{float}
\usepackage{fancyhdr}
\usepackage{titlesec}
\usepackage{xurl}
\usepackage[colorlinks=true,linkcolor=black,urlcolor=black,citecolor=black]{hyperref}
\ExplSyntaxOn
\cs_if_exist:NF \tbl_save_outer_table_cols:
  { \cs_new_protected:Npn \tbl_save_outer_table_cols: {} }
\ExplSyntaxOff
\usepackage{bidi}

\newfontfamily\arabicfont[
  Path=fonts/Amiri/,
  Extension=.ttf,
  UprightFont=Amiri-Regular,
  BoldFont=Amiri-Bold,
  ItalicFont=Amiri-Italic,
  BoldItalicFont=Amiri-BoldItalic,
  Script=Arabic,
  Language=Arabic
]{Amiri}
\newcommand{\arabictext}[1]{\RL{\arabicfont #1}}
\newcommand{\arabicprompt}[1]{\begin{RTL}\arabicfont #1\end{RTL}}

\newcolumntype{L}{>{\raggedright\arraybackslash}X}
\newcolumntype{C}{>{\centering\arraybackslash}X}
\newcolumntype{P}[1]{>{\raggedright\arraybackslash}p{#1}}
\definecolor{rowgray}{HTML}{F4F4F4}
\definecolor{rulegray}{HTML}{B8B8B8}
\definecolor{brandcolor}{HTML}{000000}

\DeclareRobustCommand{\emailaddr}[1]{\href{mailto:#1}{\nolinkurl{#1}}}
\newcommand{\organizationname}{Perle}
\newcommand{\paperdate}{\today}
\newcommand{\brandrule}{\noindent\textcolor{brandcolor}{\rule{\linewidth}{1.5pt}}\par}
\newcommand{\lightrule}{\noindent\textcolor{rulegray}{\rule{\linewidth}{0.4pt}}\par}

\titleformat{\section}{\normalfont\large\bfseries}{\thesection}{0.65em}{}
\titleformat{\subsection}{\normalfont\normalsize\bfseries}{\thesubsection}{0.65em}{}
\titlespacing{\section}{0pt}{16pt plus 3pt minus 2pt}{6pt}
\titlespacing{\subsection}{0pt}{10pt plus 2pt minus 1pt}{4pt}
\setlist[itemize]{leftmargin=1.5em,topsep=3pt,itemsep=2pt,parsep=0pt}
\setlist[enumerate]{leftmargin=1.7em,topsep=3pt,itemsep=4pt,parsep=0pt}
\newcommand{\shorttitle}{Saudi Dialect \& Cultural Competence Benchmark}
\fancypagestyle{firstpage}{
  \fancyhf{}
  
  \fancyfoot[L]{\footnotesize\textcopyright~\the\year~\organizationname. All rights reserved.}
  \fancyfoot[R]{\small 1}
}

\renewenvironment{abstract}{%
  \small\bfseries\noindent\ignorespaces
}{\par\medskip}

\begin{document}
\thispagestyle{firstpage}

\begin{flushleft}
  \IfFileExists{company_logo.png}{%
    \includegraphics[height=20pt]{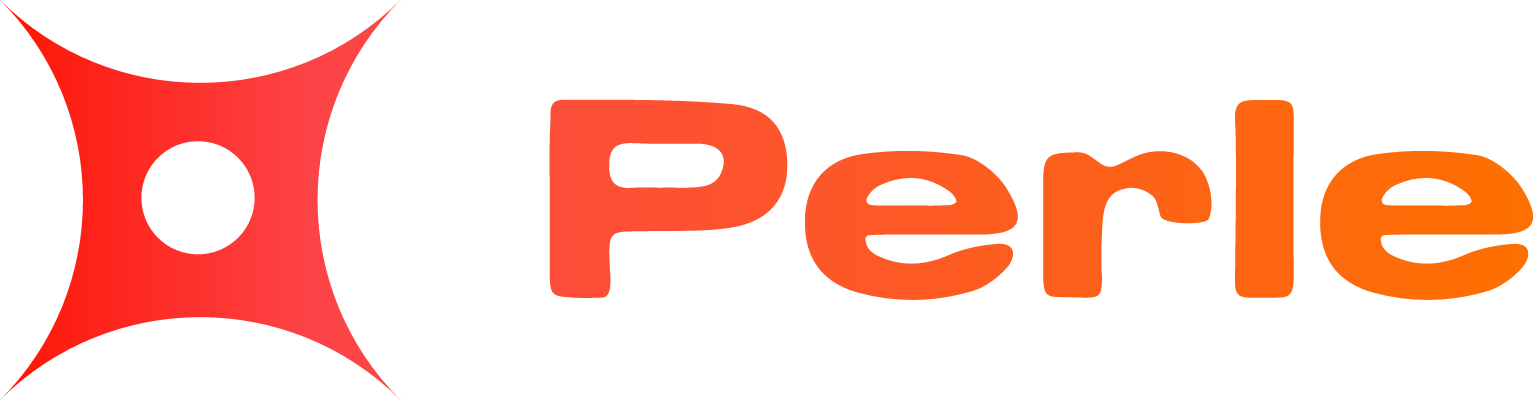}%
  }{%
    \textbf{\organizationname}%
  }
  \hfill
  {\small\paperdate}
\end{flushleft}

\vspace{4pt}
\brandrule
\vspace{8pt}
{\fontsize{18pt}{22pt}\selectfont\bfseries\noindent
Beyond Fluency: A Rubric-Based Benchmark for Evaluating Saudi Dialect and Cultural Competence in Large Language Models\par}
\vspace{4pt}
{\large\bfseries\noindent A Reference-Guided, Expert-Authored Evaluation of Four State-of-the-Art Models\par}

\vspace{12pt}

\noindent
\textbf{Ghassan Al-Sumaidaee}\textsuperscript{1,*,\dag}\footnote{Corresponding author: \emailaddr{ghassan.al-sumaidaee@perle.ai}; ORCID: \href{https://orcid.org/0000-0002-5536-0252}{0000-0002-5536-0252}},\allowbreak\quad
\textbf{Sajjad Abdoli}\textsuperscript{1,*,\dag}\footnote{Corresponding author: \emailaddr{sajjad@perle.ai}},\allowbreak\quad
\textbf{Ahmed Rashad}\textsuperscript{1},\allowbreak\quad
\textbf{Maxim Legg}\textsuperscript{1}
\par\vspace{4pt}
{\small
\textsuperscript{1}Perle\quad
\textsuperscript{*}Equal contribution.\quad
\textsuperscript{\dag}Corresponding authors\\[3pt]
\emailaddr{ghassan.al-sumaidaee@perle.ai}\quad
\emailaddr{sajjad@perle.ai}\\[2pt]
\emailaddr{ahmed@perle.ai}\quad
\emailaddr{max@perle.ai}}

\vspace{11pt}

\begin{abstract}
\noindent
Large language models are increasingly deployed in Arabic-speaking markets, yet standard benchmarks overwhelmingly reward Modern Standard Arabic (MSA) fluency while leaving dialectal and culturally-grounded competence unmeasured. This gap is consequential: everyday Arabic is largely dialectal, and dialect encodes social meaning that MSA-centric evaluation cannot capture. We present a rubric-based benchmark for the Saudi dialect, comprising 31 expert-authored prompts spanning idiomatic, pragmatic, lexical, and culturally-embedded phenomena, each paired with an expert-established ground truth. Our methodology separates evaluation into a model-agnostic phase, in which atomic, MECE positive criteria are derived solely from the ground truth, and a model-specific phase, in which four state-of-the-art systems---Claude Opus 5, Gemini 3.7, GPT-5.6, and Kimi K3---are scored against those criteria and penalised for errors they actively introduce. Across 124 model-prompt evaluations we catalogue 466 error instances under a nine-category taxonomy. The four systems cluster within a narrow macro-average band (42.7\%--53.1\%), with no model exceeding 55\% and every model recording at least one negative-scoring prompt, confirming that Saudi dialectal competence remains broadly unsolved. Notably, Ambiguous Framing is the dominant failure mode (37.3\% of errors) while outright Hallucination accounts for only 11.2\%, indicating that models fail less by stating falsehoods than by distorting register and flattening pragmatic nuance. We further observe a consistency-versus-ceiling trade-off and model-distinctive error signatures. We release the full prompt set, ground truths, and scored rubrics to support reproducible dialectal evaluation.
\end{abstract}


\vspace{6pt}
\lightrule
\vspace{14pt}

\section{Introduction}

The rapid integration of large language models (LLMs) into consumer and enterprise applications across the Arab world has outpaced our ability to measure whether these systems genuinely understand the language as it is spoken. This reflects a broader challenge in LLM benchmarking: aggregate evaluations can obscure important capability gaps across tasks, languages, and linguistic varieties \cite{hodak2023benchmarking,guo2025benchmarking}. Contemporary Arabic evaluation is dominated by Modern Standard Arabic (MSA): the formal register of news, official documents, and formal education, while culturally and sociolinguistically grounded Arabic knowledge remains much less extensively evaluated \cite{abdoli2026arabic}. MSA, however, is almost nobody's mother tongue. The Arabic that people use to greet a neighbour, tease a friend, defuse a quarrel, or ask a favour is dialectal, and it is in dialect that the culturally load-bearing meaning of an utterance most often resides. A model that scores well on MSA comprehension may still fail comprehensively at the everyday communicative tasks for which users actually reach for it.

This paper addresses that gap for one high-value variety: the Saudi dialect, together with the cultural knowledge in which it is embedded. Saudi Arabic is not a monolith---it spans Najdi, Hijazi, and other regional varieties---and its expressions frequently encode indirect social meaning: a phrase that is literally a blessing may function as a veiled threat; a word that is literally a colour may be an imperative to hurry; a request that is literally about food may be a demand for total candour. These are precisely the phenomena that expose the difference between a model that has memorised a bilingual dictionary and one that has internalised how a language is actually used.

A central difficulty in evaluating such competence is that it resists automatic scoring. There is rarely a single reference string against which a response can be matched, and open-ended model evaluation can be sensitive to both stylistic variation and prompt formulation \cite{milivcka2025benchmark,razavi2025benchmarking}. Surface-level metrics such as BLEU or exact match are therefore meaningless for open-ended pragmatic explanation. We adopt a rubric-based methodology in which subject-matter experts (SMEs)---native Saudi speakers with lived cultural knowledge---decompose the correct answer into atomic, independently-checkable criteria, and human scorers evaluate each model response against those criteria. Crucially, our design separates the construction of the evaluation standard from the observation of any model's behaviour, so that the same rubric can be applied identically to every model under test.

\subsection{Contributions}

\begin{enumerate}
  \item \textbf{A Saudi-dialect benchmark.} We release 31 expert-authored prompts, each written in authentic Saudi dialect, targeting knowledge that requires local, lived experience rather than textbook or web-searchable facts, together with an SME-validated ground truth for each.
  \item \textbf{A reproducible two-phase rubric methodology.} We formalise a model-agnostic, ground-truth-derived positive-criteria construction phase and a model-specific negative-criteria and scoring phase, ensuring that positive criteria are reused identically across all evaluated systems.
  \item \textbf{A nine-category failure taxonomy for dialectal evaluation.} We extend a six-category error scheme with categories that emerged as necessary during annotation---most notably Language Mixing---and report a full cross-model error analysis over 466 discrete error instances.
  \item \textbf{A comparative study of four state-of-the-art models.} We evaluate Claude Opus 5, Gemini 3.7, GPT-5.6, and Kimi K3, and identify model-distinctive error signatures with direct implications for deployment in Saudi-facing products.
\end{enumerate}

\subsection{Why dialectal cultural competence is hard to fake}

A well-constructed dialectal prompt has a diagnostic property: a model lacking authentic knowledge cannot fall back on fluent paraphrase. It is forced into one of two observable failure behaviours---it either hallucinates a confident but fabricated meaning, or it retreats to a superficially plausible but culturally hollow answer that misses the load-bearing nuance. Both are detectable by an SME and both are captured by our rubric. This is why prompt difficulty is validated before any rubric is built: a prompt that every model answers correctly cannot discriminate between systems and is discarded.

\section{Background and Motivation}

\subsection{Related benchmarking work}

Recent LLM evaluation has increasingly moved from broad aggregate testing toward capability-specific benchmarks. Examples include long-context modelling \cite{dong2024bamboo}, document reading systems \cite{zou2025docbench}, and the interpretation of real-world clinical text \cite{wu2026bridge}. These studies demonstrate the value of evaluation sets designed around the actual failure conditions of a particular deployment domain.

Complementary benchmarks from our group apply the same capability-specific philosophy across other modalities and language settings, including text-to-image generation \cite{abdoli2026t2i}, sound-source identification \cite{abdoli2026audio}, and multilingual code-switching speech recognition \cite{abdoli2026asr}. The present study focuses this approach on Saudi dialect and culture, extending earlier evaluation of Arabic cultural and sociolinguistic knowledge \cite{abdoli2026arabic} with a shared-positive, model-specific-negative rubric design and a detailed error taxonomy.

\subsection{The MSA bias in Arabic evaluation}

Arabic is a canonical example of diglossia: a high (formal, written) variety coexists with a family of low (spoken, regional) varieties that differ substantially in phonology, morphology, and lexicon. Because annotated resources are far more abundant for MSA, evaluation practice has gravitated toward it. Related evaluations of Arabic cultural knowledge and multilingual code-switching likewise show why performance on formal or monolingual inputs cannot be assumed to transfer to socially situated language use \cite{abdoli2026arabic,abdoli2026asr}. The result is a systematic blind spot: models can appear competent in Arabic while being unable to correctly interpret common spoken expressions. For applications ranging from customer support to conversational assistants, this blind spot is exactly where user experience is won or lost.

\subsection{Cultural knowledge as an evaluation target}

Beyond vocabulary, dialect carries cultural knowledge transmitted through lived experience rather than formal instruction---proverbs and their origins, etiquette for answering a call, indirect conflict-resolution norms, and material-culture references such as traditional footwear. This knowledge is difficult to acquire from web text at scale and is therefore an especially demanding test of whether a model has genuinely absorbed a community's communicative practices. Our benchmark deliberately concentrates on this stratum.

\subsection{Rubric-based human evaluation}

When a task has no single reference answer, decomposed rubric scoring offers a disciplined alternative to holistic rating. By breaking the ground truth into atomic criteria, each independently marked pass or fail, rubric scoring improves inter-rater reliability, localises exactly where a model succeeds or fails, and yields interpretable, actionable diagnostics rather than an opaque scalar. Our methodology operationalises two principles---atomicity and MECE coverage---to make this decomposition rigorous and repeatable.

\section{Benchmark Construction}

The benchmark comprises 31 prompts authored by native Saudi SMEs. Each prompt satisfies a set of design constraints intended to maximise diagnostic value, and each is classified along two axes---evaluation scope and linguistic phenomenon---that structure the later analysis.

\subsection{Prompt design principles}

Every prompt was required to satisfy the following criteria, enforced at authoring time:

\begin{itemize}
  \item \textbf{Local-knowledge requirement.} The correct answer should require lived, in-community experience; anything resolvable by a routine web search or a general Arabic textbook is too easy to be discriminative.
  \item \textbf{Written in the target dialect.} Prompts are composed in Saudi dialect (Najdi/Hijazi as appropriate), not MSA, so that register itself is part of what is tested and the model is not cued toward a formal response.
  \item \textbf{Conversational and natural.} Prompts read like a real person asking a genuine question, testing whether the model can engage in-register rather than defaulting to an encyclopaedic style.
  \item \textbf{Multi-layered where possible.} Prompts invite a layered answer---meaning, usage, and cultural grounding---so a single response can be scored along several dimensions.
  \item \textbf{Non-leading.} Prompts avoid embedding the answer or vocabulary the model can echo back.
  \item \textbf{Culturally sensitive, not offensive.} Sensitive themes may appear but are framed respectfully; the goal is to test knowledge, not to provoke harmful output.
\end{itemize}

\subsection{Scope: cultural vs. linguistic}

Each prompt is assigned a single primary scope. Linguistic prompts test identification, use, and explanation of dialect-specific vocabulary, idioms, proverbs, and speech patterns. Cultural prompts test knowledge of traditions, social norms, material culture, and collective memory. A prompt that touches both is assigned to its dominant dimension rather than mixed. In the present set the large majority of prompts are linguistic, reflecting the density of idiomatic and pragmatic traps in spoken Saudi Arabic, with a minority classified as cultural (for example, the identification of traditional men's footwear, and the reading of indirect post-conflict speech).

\subsection{Difficulty validation}

Before any rubric was constructed, each candidate prompt was submitted to a model in a clean, memory-disabled session as a quick discriminativeness check. Prompts that were answered perfectly were revised to target deeper local knowledge or discarded. This validation output was never used for scoring; it served only to confirm that the prompt could expose a knowledge gap. This step protects annotation effort from being spent on non-discriminative items.

\subsection{Illustrative prompts}

Table~\ref{tab:illustrative-prompts} gives representative examples spanning the phenomena in the benchmark, with the load-bearing element that a correct answer must capture and the literal misreading that constitutes the characteristic trap.

\begin{table}[H]
\centering
\caption{Representative benchmark prompts, the culturally load-bearing meaning each targets, and the literal misreading that defines its failure mode.}
\label{tab:illustrative-prompts}
\small
\begin{tabularx}{\textwidth}{P{0.22\textwidth} P{0.17\textwidth} L L}
\toprule
\textbf{Prompt (Arabic)} & \textbf{Phenomenon} & \textbf{Intended meaning} & \textbf{Characteristic trap} \\
\midrule
\rowcolor{rowgray} \arabictext{الله يبيض وجهك} & Idiom of gratitude & A high compliment / thanks rooted in honour & Literal ``may God whiten your face'' (skin colour) \\
\arabictext{أبشر} & Pragmatic reply & ``Consider it done / at your service'' & Literal ``rejoice / good news'' \\
\rowcolor{rowgray} \arabictext{فلان سنع} & Evaluative lexeme & Well-mannered + socially intelligent & Flattened to generic ``good/nice'' \\
\arabictext{أزرق البقالة وتعال} & Dialectal verb & ``Dash to the shop and come back'' & Literal ``blue the grocery'' \\
\rowcolor{rowgray} \arabictext{طقها والحقها} & Idiom & Acts first, deals with fallout later & Literal ``hit it and catch it'' \\
\arabictext{لا تبيع الموية في حارة السقايين} & Proverb & Don't sell to those who already have it & Read flatly, missing the metaphor \\
\bottomrule
\end{tabularx}
\end{table}

\section{Evaluation Methodology}

Our methodology is deliberately split into two phases so that the evaluation standard is fixed before any model output is observed. This separation is the mechanism that makes cross-model comparison fair: every model is graded against an identical, pre-committed set of positive criteria.

\subsection{Two-phase workflow}

\textbf{Phase A (pre-output, model-agnostic).} The SME establishes the ground truth, decomposes it into atomic positive criteria, assigns each a dimension and weight, and finalises the prompt. No model output is consulted. The deliverable is a reusable prompt-plus-positive-rubric pair.

\textbf{Phase B (post-output, model-specific).} For each model, the response is generated in a clean session, scored pass/fail against every positive criterion, and then examined for errors it actively introduced. Each such error becomes a negative criterion carrying a penalty. Because different models make different mistakes, the negative criteria are model-specific, whereas the positive criteria are shared.

\subsection{Criterion design: atomicity and MECE}

Two principles govern criterion construction. Atomicity requires that each criterion test exactly one thing, so that a response cannot be simultaneously right and wrong under a single criterion. MECE (mutually exclusive, collectively exhaustive) requires that no two criteria overlap and that together they cover the complete ground truth. Together these principles ensure that the total score is a faithful, non-double-counted reflection of how much of the correct answer a model produced.

\subsection{Scoring fields}

Each criterion---positive or negative---carries the following fields:

\begin{itemize}
  \item \textbf{Dimension:} Accuracy, Context Awareness, Completeness, or Communication Quality.
  \item \textbf{Objectivity:} whether the criterion has a clear right/wrong answer or requires judgement (e.g., naturalness of register).
  \item \textbf{Explicitness:} whether the target must be stated explicitly or may be conveyed implicitly.
  \item \textbf{Weight:} the point value; for positive criteria this equals the maximum score, for negative criteria it is negative.
  \item \textbf{Score type and max score:} positive criteria are scored against their weight; negative criteria have a maximum of 0, since the best outcome is that the error is absent.
  \item \textbf{Result and justification:} pass/fail (or error-present/absent), with a justification referencing specific response content; failures and negative criteria carry a bracketed error-type tag.
\end{itemize}

\subsection{Scoring model and equations}

Let a prompt have positive criteria indexed by $i$ with weights $w_i$, and let the model under evaluation trigger negative criteria indexed by $j$ with penalties $p_j$ ($p_j>0$ denotes the magnitude deducted). Define the pass indicator $x_i\in\{0,1\}$ and the error indicator $e_j\in\{0,1\}$. The maximum attainable score for a prompt depends only on its positive criteria:
\begin{equation} S_{\max}=\sum_i w_i. \end{equation}
The points a model earns from positive criteria, and the penalty it incurs from its active errors, are:
\begin{align}
S_{\mathrm{pos}}&=\sum_i w_i x_i,\\
P&=\sum_j p_j e_j.
\end{align}
The prompt-level model score and its normalised percentage are then:
\begin{align}
S&=S_{\mathrm{pos}}-P,\\
\mathrm{Score\%}&=100\times\frac{S}{S_{\max}}.
\end{align}
Because penalties are not bounded by $S_{\mathrm{pos}}$, a prompt-level score can be negative. This is intentional: a response that supplies a wrong answer and compounds it with fabrications is actively misleading, not merely unhelpful, and the scoring model reflects that a negative outcome is a meaningful and valid result. A model's aggregate score over the benchmark of $N$ prompts is the mean of its per-prompt percentages,
\begin{equation}
\mathrm{Score\%}_{\mathrm{agg}}=\frac{1}{N}\sum_k \mathrm{Score\%}_k.
\end{equation}

\subsection{Controlled generation conditions}

All model outputs were captured under uniform conditions to prevent contamination: memory and conversation history disabled, one prompt per session, the first response taken verbatim with no re-prompting, and formatting preserved exactly as emitted. These controls ensure that each response reflects the model's independent behaviour on the prompt rather than an artefact of dialogue history.

\subsection{Workflow summary}

\begin{table}[H]
\centering
\caption{The six-step, two-phase evaluation workflow. Steps 1--4 are performed once per prompt; steps 5--6 are repeated for every model.}
\label{tab:workflow}
\small
\begin{tabularx}{\textwidth}{c c P{0.27\textwidth} L}
\toprule
\textbf{Step} & \textbf{Phase} & \textbf{Action} & \textbf{Deliverable} \\
\midrule
\rowcolor{rowgray} 1 & A & Define scope & Cultural or linguistic (never mixed) \\
2 & A & Draft prompt & Dialect-native, locally-rooted, multi-layered \\
\rowcolor{rowgray} 3 & A & Validate difficulty & Confirm the prompt is discriminative \\
4 & A & Build positive criteria & Atomic, MECE, weighted (model-agnostic) \\
\rowcolor{rowgray} 5 & B & Generate output & Clean session, verbatim capture \\
6 & B & Score \& add negatives & Pass/fail, penalties, justifications, final score \\
\bottomrule
\end{tabularx}
\end{table}

\section{Failure Taxonomy}

Every failed positive criterion and every negative criterion is tagged with an error type. We began from a six-category scheme and extended it during annotation as recurring behaviours emerged that the original categories did not cleanly capture. The final nine-category taxonomy is given in Table~\ref{tab:taxonomy}.

\begin{table}[H]
\centering
\caption{The nine-category failure taxonomy. The final two categories---Formatting and Language Mixing---were added during annotation; Instruction Violation was promoted to a first-class category to capture explicit-request failures distinct from mere omission.}
\label{tab:taxonomy}
\small
\begin{tabularx}{\textwidth}{P{0.24\textwidth} L}
\toprule
\textbf{Error type} & \textbf{Definition} \\
\midrule
\rowcolor{rowgray} Hallucination & Fabricates information: invents a non-existent term, a false etymology, or a fictional origin, or states something confidently and incorrectly. \\
Omission & Fails to mention a critical element the correct answer requires; the point is not addressed at all. \\
\rowcolor{rowgray} Ambiguous Framing & Presents information misleadingly or unclearly without being outright false---including wrong register, inappropriate mixing of formal and informal, or framing that distorts significance. \\
Cultural Misattribution & Attributes a term, practice, or tradition to the wrong region, community, or referent. \\
\rowcolor{rowgray} Irrelevant Addition & Adds unsolicited content that does not serve the question. \\
Rendering Failure & Produces garbled text, invalid characters, or broken Arabic script---a technical rather than a knowledge failure. \\
\rowcolor{rowgray} Formatting & Presentation defects that impede readability or violate the expected response shape (e.g., inappropriate structure or markup artefacts). \\
Instruction Violation & Fails to do what the prompt explicitly asked---e.g., answering a different question, asking the user a question instead of answering, or ignoring an explicit format request. \\
\rowcolor{rowgray} Language Mixing & Introduces the wrong language or variety inappropriately---e.g., drifting into MSA or English where Saudi dialect was required, or code-switching that breaks register. \\
\bottomrule
\end{tabularx}
\end{table}

Two taxonomy notes bear on interpretation. First, Ambiguous Framing is intentionally broad and is the category most reliant on evaluator judgement; because it absorbs register and nuance-distortion failures that are central to dialectal competence, it is both the most informative and the most in need of inter-rater scrutiny. Second, Language Mixing overlaps conceptually with Ambiguous Framing (a wrong-register response can be seen as both); we treat an inappropriate switch of language or variety as Language Mixing and reserve Ambiguous Framing for distortions within the correct variety, but this boundary is a modelling choice that future work may refine.

\section{Quantitative Performance}

We complement the error analysis with the numeric rubric scores. Each prompt yields a per-model percentage via Eq.~(5); a model's aggregate is summarised both as a macro-average (the unweighted mean of its 31 per-prompt percentages, Eq.~(6)) and a micro-average (total points earned divided by total points available, which weights each prompt by its maximum score). Reporting both guards against a small number of high-max prompts dominating the headline number.

\subsection{Leaderboard}

Table~\ref{tab:leaderboard} presents the leaderboard. The four systems cluster within a narrow $\sim$10-point macro-average band (42.7\%--53.1\%), underscoring that Saudi-dialect cultural competence remains an unsolved and broadly difficult task: no model reaches 55\%, and every model records at least one negative-scoring prompt where active errors outweighed the points earned.

\begin{table}[H]
\centering
\caption{Leaderboard over 31 prompts. Macro \% is the mean of per-prompt percentages; Micro \% weights by prompt max score; SD is the standard deviation of per-prompt percentages; ``$\geq$50\%'' counts prompts scoring at least half; ``Neg.'' counts prompts with a negative score. Ranked by macro-average.}
\label{tab:leaderboard}
\small
\begin{tabular}{lrrrrrrrr}
\toprule
\textbf{Model} & \textbf{Macro \%} & \textbf{SD} & \textbf{Micro \%} & \textbf{Median} & \textbf{Min} & \textbf{Max} & \textbf{$\geq$50\%} & \textbf{Neg.} \\
\midrule
\rowcolor{rowgray} Kimi K3 & 53.1 & 22.7 & 52.9 & 58.3 & $-21.2$ & 80.6 & 22/31 & 2 \\
GPT-5.6 & 52.6 & 33.7 & 54.3 & 61.5 & $-44.8$ & 88.9 & 24/31 & 3 \\
\rowcolor{rowgray} Gemini 3.7 & 50.7 & 16.6 & 51.5 & 54.2 & $-6.2$ & 79.4 & 19/31 & 1 \\
Claude Opus 5 & 42.7 & 29.2 & 43.0 & 55.2 & $-24.2$ & 76.5 & 20/31 & 4 \\
\bottomrule
\end{tabular}
\end{table}

\begin{figure}[H]
\centering
\includegraphics[width=0.82\textwidth]{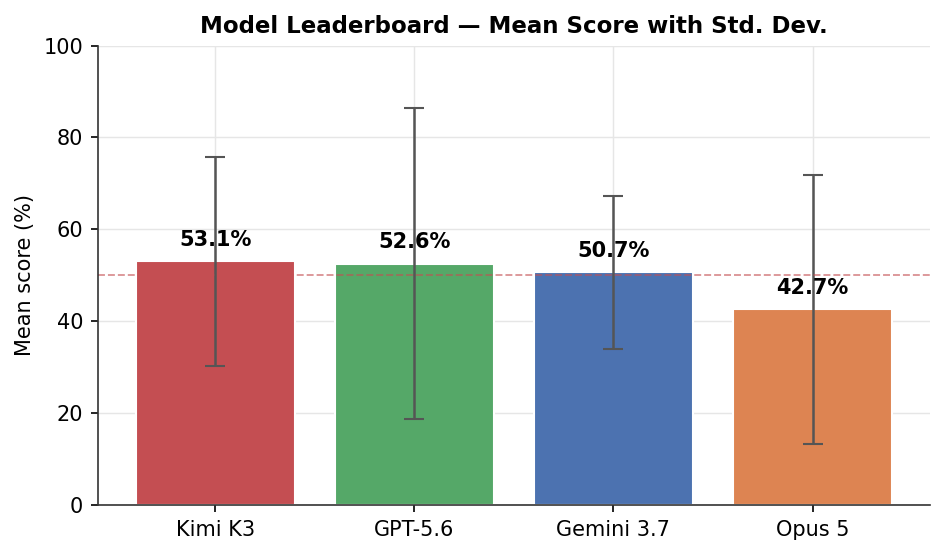}
\caption{Mean score per model with standard-deviation error bars. The dashed line marks the 50\% threshold. Differences among the top three models are small relative to their spread.}
\label{fig:leaderboard}
\end{figure}

\subsection{Consistency versus peak performance}

\begin{figure}[H]
\centering
\includegraphics[width=0.88\textwidth]{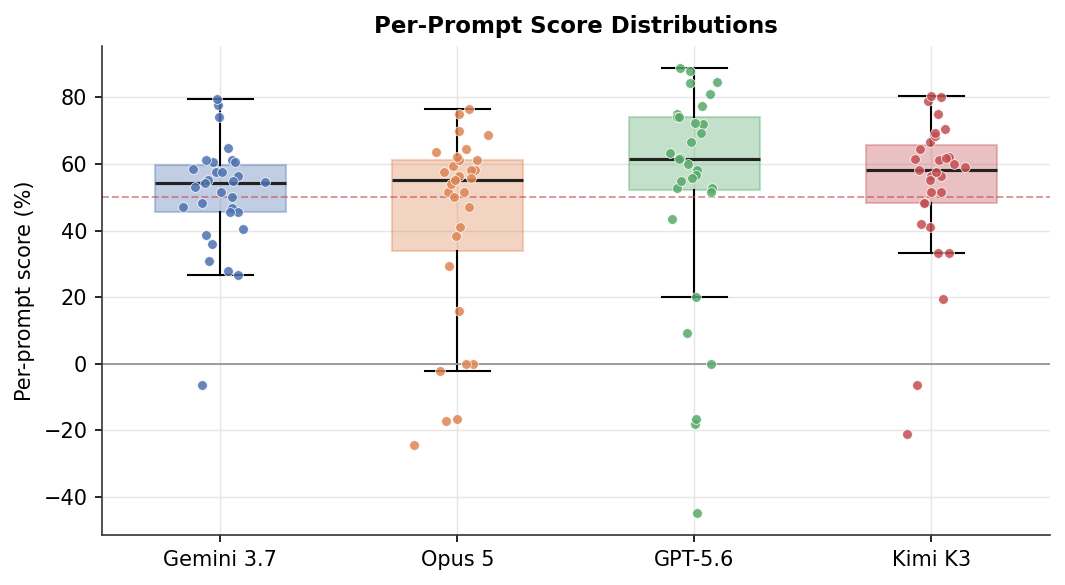}
\caption{Per-prompt score distributions (box plots with individual prompts overlaid). Solid line at 0\%, dashed line at 50\%. GPT-5.6 has the highest median but the greatest spread; Gemini 3.7 is tightest.}
\label{fig:distributions}
\end{figure}

Aggregate means conceal a sharp difference in reliability. Figure~\ref{fig:distributions} shows the full per-prompt distributions. GPT-5.6 attains the highest median (61.5\%) and the highest single-prompt score (88.9\%) but also the widest dispersion (SD 33.7) and the worst single failure ($-44.8$\% on SAU-4). Gemini 3.7, by contrast, is the most consistent system (SD 16.6) with only one negative prompt, despite a lower ceiling. Claude Opus 5 combines a moderate median with heavy left-tail risk (four negative prompts, SD 29.2). This consistency--ceiling trade-off is decision-relevant: a deployment that cannot tolerate occasional severe failures may prefer the tighter distribution even at a lower mean.

\subsection{Macro versus micro aggregation}

\begin{figure}[H]
\centering
\includegraphics[width=0.82\textwidth]{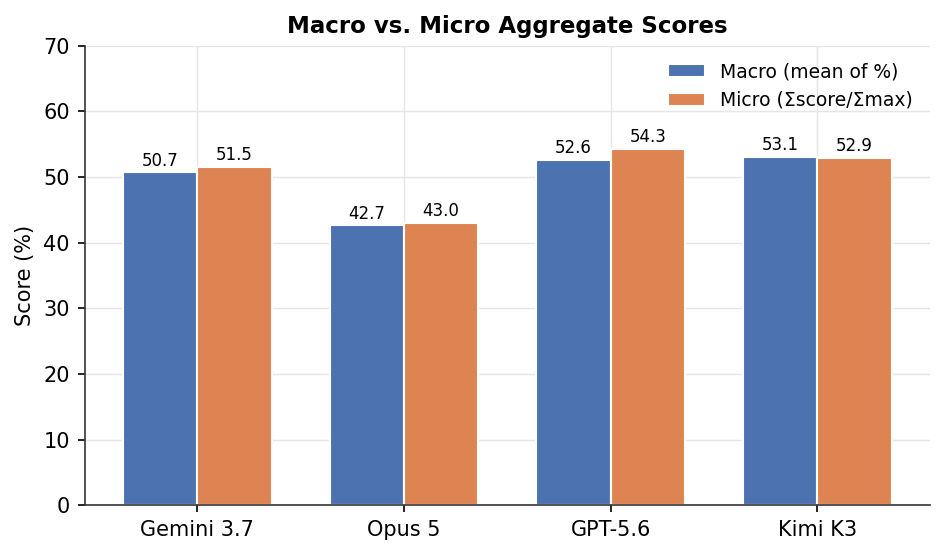}
\caption{Macro- versus micro-average scores. Ranking is invariant to the choice of aggregation; the small gaps confirm robustness.}
\label{fig:macro-micro}
\end{figure}

Figure~\ref{fig:macro-micro} contrasts the two aggregation schemes. The rankings are stable across schemes and the macro/micro gap is small for every model (at most $\sim$1.6 points), indicating that no model's standing is an artefact of a few high-weight prompts. The largest divergence is for GPT-5.6, whose micro-average slightly exceeds its macro-average because it performed well on several high-max prompts (e.g., SAU-25, max 58).

\subsection{Item difficulty: which prompts were hardest}

\begin{figure}[H]
\centering
\includegraphics[width=\textwidth]{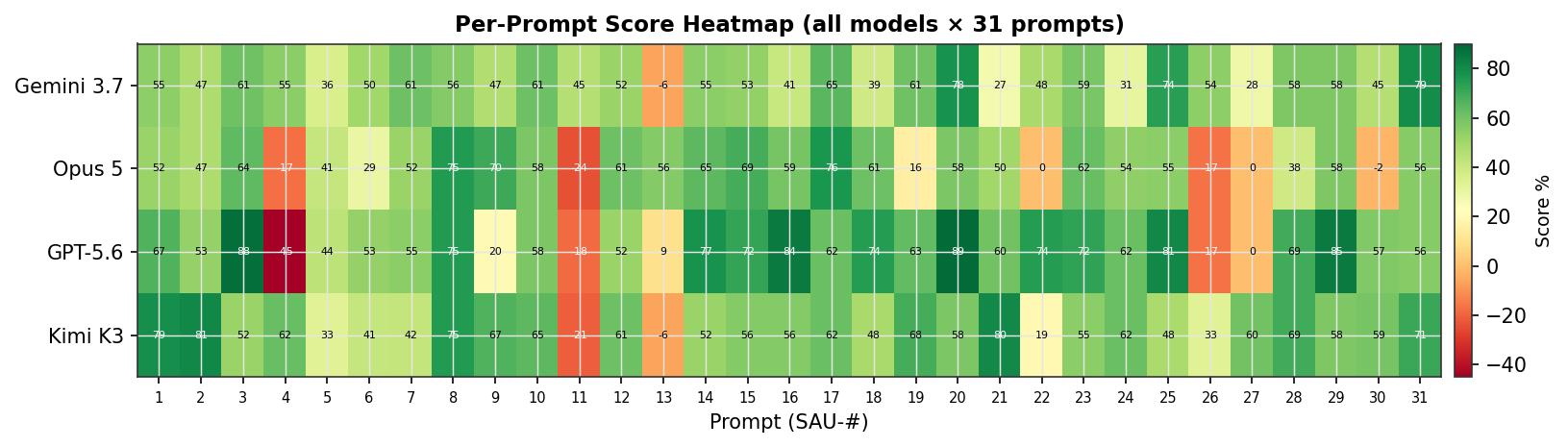}
\caption{Per-prompt score heatmap (models $\times$ 31 prompts). Green is high, red is low/negative. Vertical cool bands (e.g., SAU-11) are prompts that were hard for every model; warmer rows indicate broadly stronger models.}
\label{fig:prompt-heatmap}
\end{figure}

Averaging each prompt across the four models yields an item-difficulty ranking. Tables~\ref{tab:hardest} and~\ref{tab:easiest} list the five hardest and five easiest prompts. The hardest items concentrate on highly localised lexis and metaphor where a literal misreading is both tempting and catastrophic---for example SAU-11 (the Hijazi metaphor ``kubb al-`asha'') and SAU-27 (the nuance-laden ``dalakh''). The easiest items are common greetings and high-frequency expressions. The per-prompt heatmap (Figure~\ref{fig:prompt-heatmap}) visualises this structure: vertical cool bands mark prompts that defeated all models, while broad warm rows mark models that were broadly competent.

\begin{table}[H]
\centering
\caption{The five hardest prompts by cross-model mean score. Per-model values are percentages.}
\label{tab:hardest}
\small
\begin{tabular}{lrrrrr}
\toprule
\textbf{Prompt} & \textbf{Cross-model mean} & \textbf{Gemini} & \textbf{Opus 5} & \textbf{GPT-5.6} & \textbf{Kimi K3} \\
\midrule
\rowcolor{rowgray} SAU-11 & $-4.5$\% & 45 & $-24$ & $-18$ & $-21$ \\
SAU-13 & 13.3\% & $-6$ & 56 & 9 & $-6$ \\
\rowcolor{rowgray} SAU-26 & 13.5\% & 54 & $-17$ & $-17$ & 33 \\
SAU-4 & 13.8\% & 55 & $-17$ & $-45$ & 62 \\
\rowcolor{rowgray} SAU-27 & 22.0\% & 28 & 0 & 0 & 60 \\
\bottomrule
\end{tabular}
\end{table}

\begin{table}[H]
\centering
\caption{The five easiest prompts by cross-model mean score.}
\label{tab:easiest}
\small
\begin{tabular}{lrrrrr}
\toprule
\textbf{Prompt} & \textbf{Cross-model mean} & \textbf{Gemini} & \textbf{Opus 5} & \textbf{GPT-5.6} & \textbf{Kimi K3} \\
\midrule
\rowcolor{rowgray} SAU-20 & 70.8\% & 78 & 58 & 89 & 58 \\
SAU-8 & 70.3\% & 56 & 75 & 75 & 75 \\
\rowcolor{rowgray} SAU-17 & 66.2\% & 65 & 76 & 62 & 62 \\
SAU-3 & 65.9\% & 61 & 64 & 88 & 52 \\
\rowcolor{rowgray} SAU-31 & 65.4\% & 79 & 56 & 56 & 71 \\
\bottomrule
\end{tabular}
\end{table}

\subsection{Score--error relationship}

The quantitative scores and the error taxonomy are two views of the same evaluations. The negative-scoring prompts are precisely those where high-penalty errors---typically Hallucination or Cultural Misattribution on a load-bearing criterion---overwhelmed the positive points earned. This is why total error count (Section~\ref{sec:error-analysis}) and mean score do not rank the models identically: Claude Opus 5 has the highest error instance count and also the lowest macro-average, but GPT-5.6, with the fewest errors, does not top the leaderboard because its errors, when they occur, tend to fall on high-max prompts and include severe single failures. Error frequency and error severity are distinct axes, and a complete evaluation requires both.

\section{Error Analysis}
\label{sec:error-analysis}

We report results over 124 model-prompt evaluations (31 prompts $\times$ 4 models). Across these, human scorers recorded 466 discrete error instances. This section first presents the aggregate error load, then the cross-model comparison, and finally the model-distinctive error signatures.

\subsection{Aggregate error load}

Table~\ref{tab:model-errors} gives the per-model total error counts. GPT-5.6 accrued the fewest total error instances (103) and Claude Opus 5 the most (132), with Gemini 3.7 (117) and Kimi K3 (114) in between. Total error count is a coarse proxy---a single high-weight error can matter more than several minor ones---but it establishes the overall reliability ordering at the instance level.

\begin{table}[H]
\centering
\caption{Total error instances per model across 31 prompts, with each model's share of the 466-instance corpus.}
\label{tab:model-errors}
\small
\begin{tabular}{lrr}
\toprule
\textbf{Model} & \textbf{Total errors} & \textbf{Share of all errors} \\
\midrule
\rowcolor{rowgray} GPT-5.6 & 103 & 22.1\% \\
Kimi K3 & 114 & 24.5\% \\
\rowcolor{rowgray} Gemini 3.7 & 117 & 25.1\% \\
Claude Opus 5 & 132 & 28.3\% \\
\bottomrule
\end{tabular}
\end{table}

\begin{figure}[H]
\centering
\includegraphics[width=0.82\textwidth]{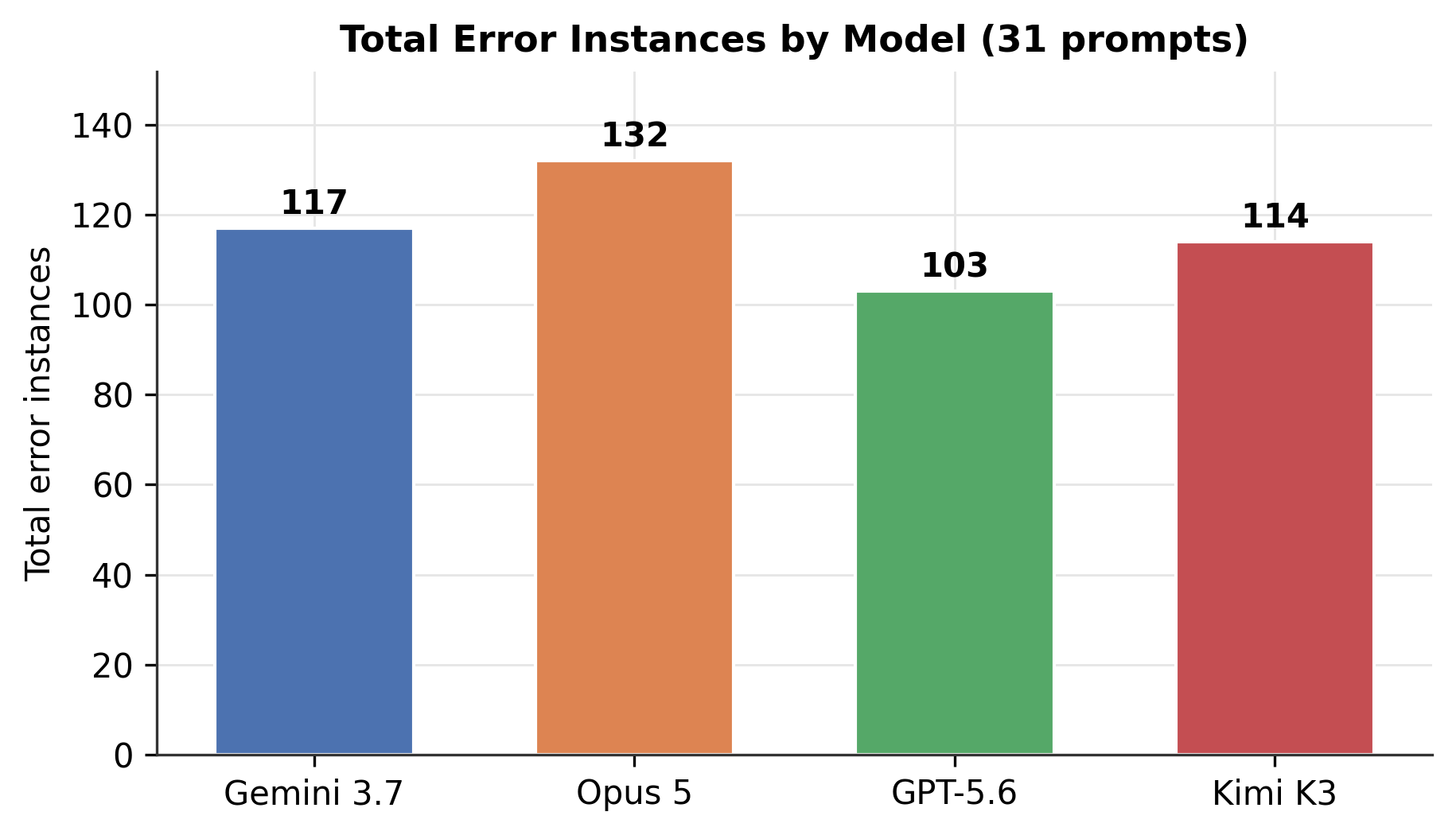}
\caption{Total error instances by model. GPT-5.6 exhibits the lowest instance-level error load; Claude Opus 5 the highest, driven substantially by non-accuracy categories (Section~\ref{sec:cross-model-errors}).}
\label{fig:total-errors}
\end{figure}

\subsection{The dominant failure mode is framing, not fabrication}

\begin{figure}[H]
\centering
\includegraphics[width=0.88\textwidth]{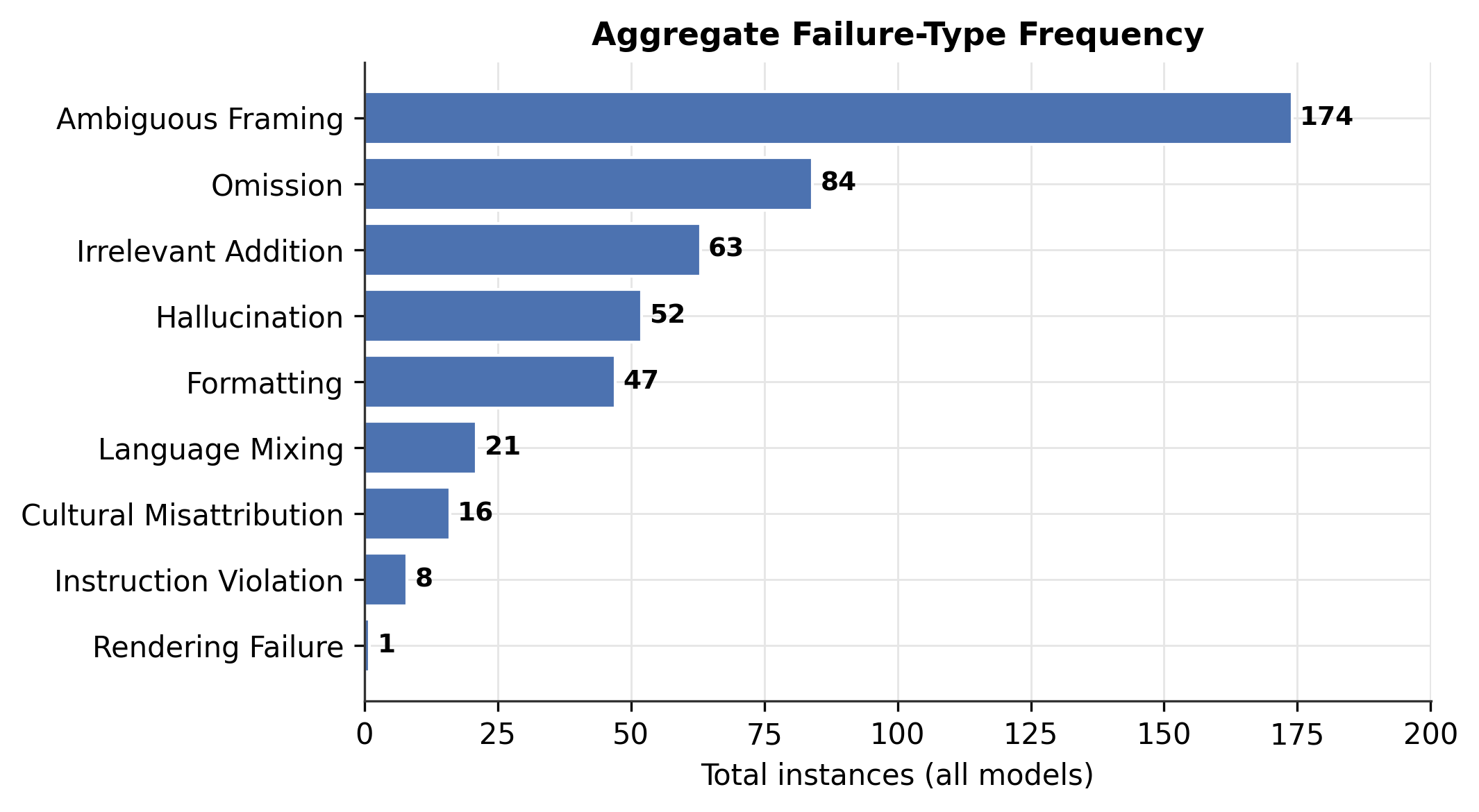}
\caption{Aggregate failure-type frequency across all models. The concentration in Ambiguous Framing, Omission, and Irrelevant Addition indicates that current models fail on Saudi-dialect tasks primarily through nuance distortion and calibration of scope, not through overt factual fabrication.}
\label{fig:aggregate-failures}
\end{figure}

The most consequential finding of the study is visible in the aggregate distribution (Figure~\ref{fig:aggregate-failures}, Table~\ref{tab:aggregate-failures}): the modal failure is not the fabrication of false facts but the subtle mishandling of framing and completeness. Ambiguous Framing alone accounts for 174 of 466 instances (37.3\%). Omission (84; 18.0\%) and Irrelevant Addition (63; 13.5\%) follow. Hallucination---often assumed to be the primary risk of LLMs on knowledge-intensive tasks---is only the fourth most common category (52; 11.2\%).

\begin{table}[H]
\centering
\caption{Aggregate failure-type frequency across all four models (466 total instances), sorted by frequency.}
\label{tab:aggregate-failures}
\small
\begin{tabular}{lrr}
\toprule
\textbf{Error type} & \textbf{Total} & \textbf{\% of all errors} \\
\midrule
\rowcolor{rowgray} Ambiguous Framing & 174 & 37.3\% \\
Omission & 84 & 18.0\% \\
\rowcolor{rowgray} Irrelevant Addition & 63 & 13.5\% \\
Hallucination & 52 & 11.2\% \\
\rowcolor{rowgray} Formatting & 47 & 10.1\% \\
Language Mixing & 21 & 4.5\% \\
\rowcolor{rowgray} Cultural Misattribution & 16 & 3.4\% \\
Instruction Violation & 8 & 1.7\% \\
\rowcolor{rowgray} Rendering Failure & 1 & 0.2\% \\
\bottomrule
\end{tabular}
\end{table}

The interpretive weight of this result is substantial. On dialectal-cultural tasks, the failures that matter are those a fluent-sounding model is most likely to commit unnoticed: giving a technically-related but register-wrong rendering, omitting the one culturally load-bearing element, or burying the answer under unsolicited elaboration. These are exactly the errors that automated fluency metrics cannot detect and that only reference-guided human scoring surfaces.

\subsection{Cross-model comparison by error type}
\label{sec:cross-model-errors}

\begin{figure}[H]
\centering
\includegraphics[width=\textwidth]{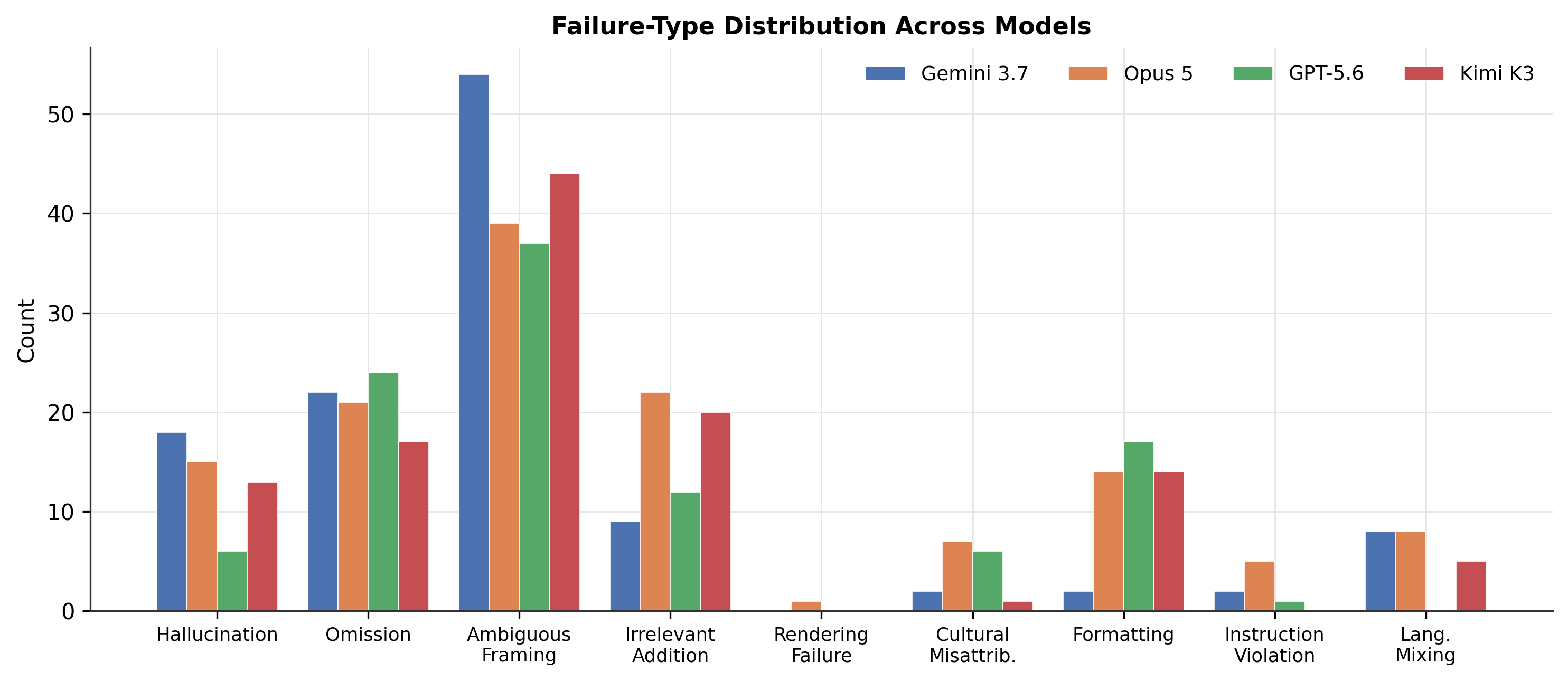}
\caption{Failure-type distribution across models (grouped). Bars are absolute instance counts. Note the model-specific concentrations: Gemini in framing, GPT-5.6 in formatting with minimal hallucination, and Opus 5 in irrelevant additions.}
\label{fig:failure-by-model}
\end{figure}

Figure~\ref{fig:failure-by-model} and Table~\ref{tab:error-crosstab} break the counts down by model and error type. Several patterns stand out. Ambiguous Framing is the leading category for every model but is most concentrated in Gemini 3.7 (54 instances; 46.2\% of its errors). GPT-5.6 has by far the lowest Hallucination count (6) yet the highest Formatting load (17). Claude Opus 5 shows the most even spread and the highest Irrelevant Addition count (22), together with the highest Cultural Misattribution (7) and Instruction Violation (5) counts. Kimi K3 resembles Gemini in its framing-heavy profile but with more Irrelevant Addition.

\begin{table}[H]
\centering
\caption{Full cross-tabulation of failure-type counts by model. The final row gives per-model totals. \texttt{LANG\_MIXING} labels and the misspelled Instruction-Violation variants in the source sheets were normalised to the canonical categories of Table~\ref{tab:taxonomy}.}
\label{tab:error-crosstab}
\small
\begin{tabular}{lrrrr}
\toprule
\textbf{Error type} & \textbf{Gemini 3.7} & \textbf{Opus 5} & \textbf{GPT-5.6} & \textbf{Kimi K3} \\
\midrule
\rowcolor{rowgray} Hallucination & 18 & 15 & 6 & 13 \\
Omission & 22 & 21 & 24 & 17 \\
\rowcolor{rowgray} Ambiguous Framing & 54 & 39 & 37 & 44 \\
Irrelevant Addition & 9 & 22 & 12 & 20 \\
\rowcolor{rowgray} Rendering Failure & 0 & 1 & 0 & 0 \\
Cultural Misattribution & 2 & 7 & 6 & 1 \\
\rowcolor{rowgray} Formatting & 2 & 14 & 17 & 14 \\
Instruction Violation & 2 & 5 & 1 & 0 \\
\rowcolor{rowgray} Language Mixing & 8 & 8 & 0 & 5 \\
\midrule
\textbf{Total} & \textbf{117} & \textbf{132} & \textbf{103} & \textbf{114} \\
\bottomrule
\end{tabular}
\end{table}

\begin{figure}[H]
\centering
\includegraphics[width=\textwidth]{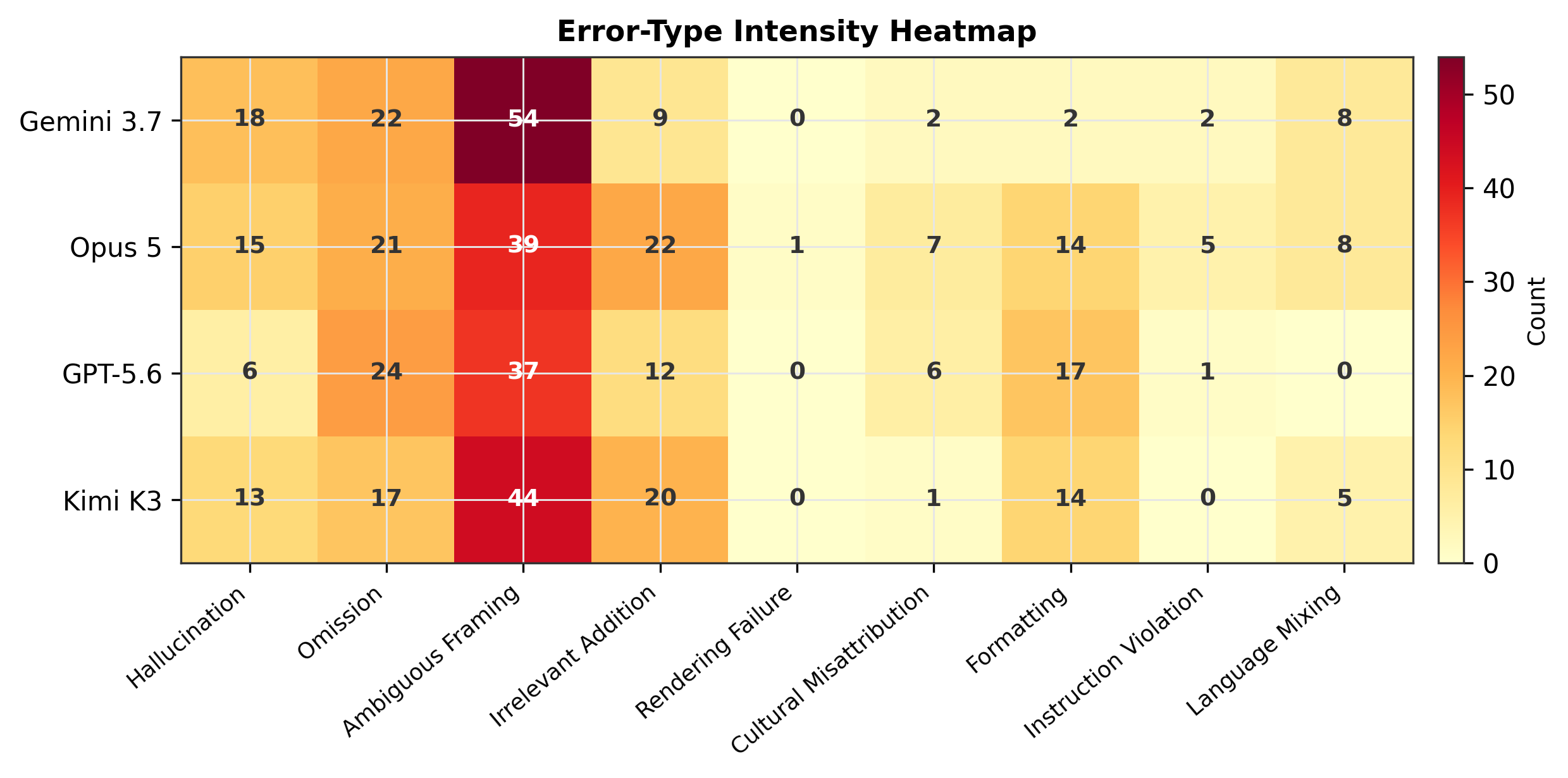}
\caption{Error-type intensity heatmap. Colour encodes count. The heatmap makes the model signatures legible at a glance: the framing column dominates, GPT-5.6's near-empty hallucination cell contrasts with its dark formatting cell, and Opus 5's irrelevant-addition cell is the darkest in its row-group.}
\label{fig:error-heatmap}
\end{figure}

\subsection{Normalised error signatures}

\begin{figure}[H]
\centering
\includegraphics[width=0.9\textwidth]{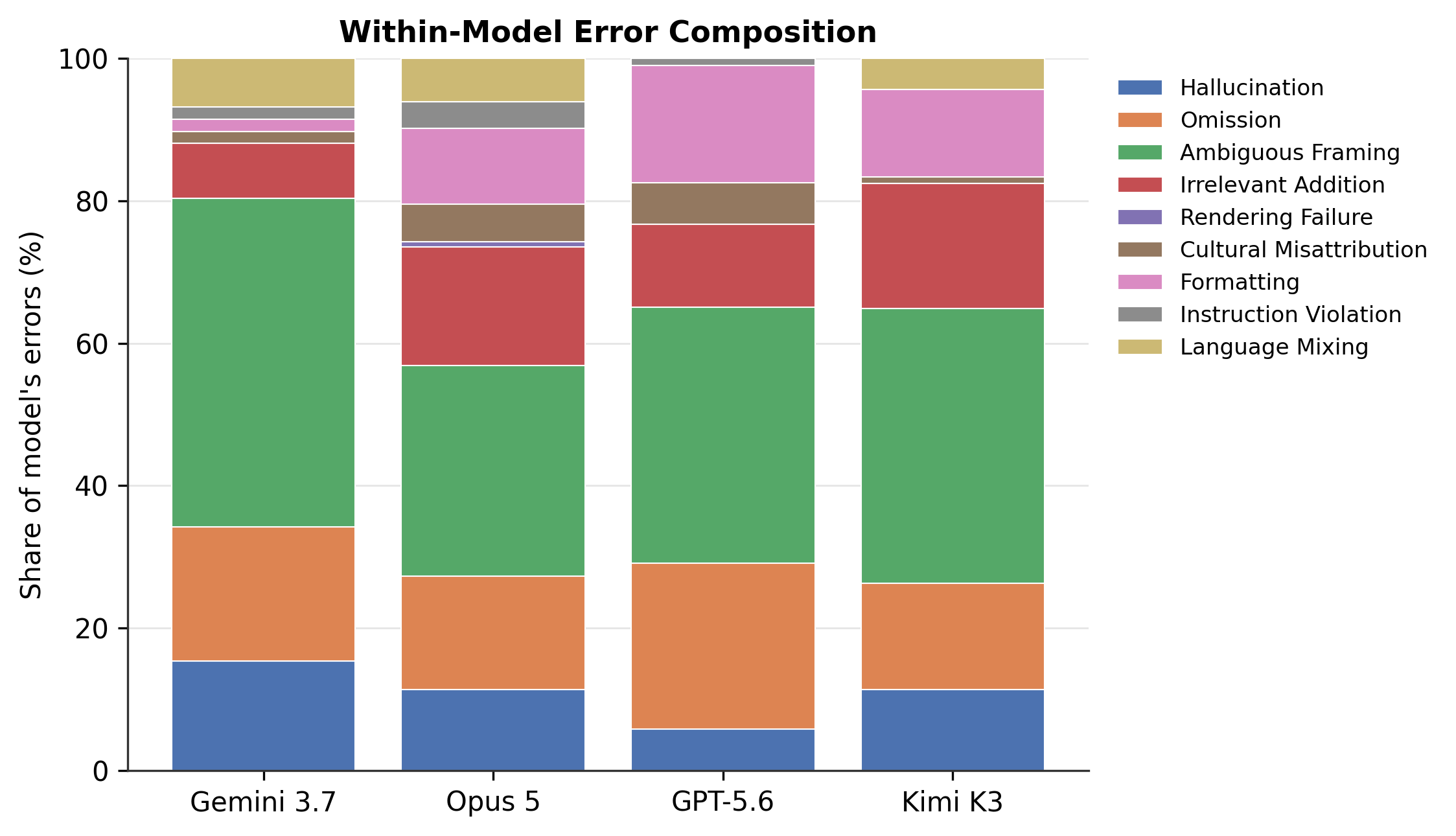}
\caption{Within-model error composition (each bar sums to 100\%). GPT-5.6 devotes the largest share of its errors to Omission and Formatting and the smallest to Hallucination; Gemini devotes nearly half its errors to Ambiguous Framing; Opus 5 and Kimi K3 carry the heaviest Irrelevant-Addition shares.}
\label{fig:error-composition}
\end{figure}

\begin{figure}[H]
\centering
\includegraphics[width=0.82\textwidth]{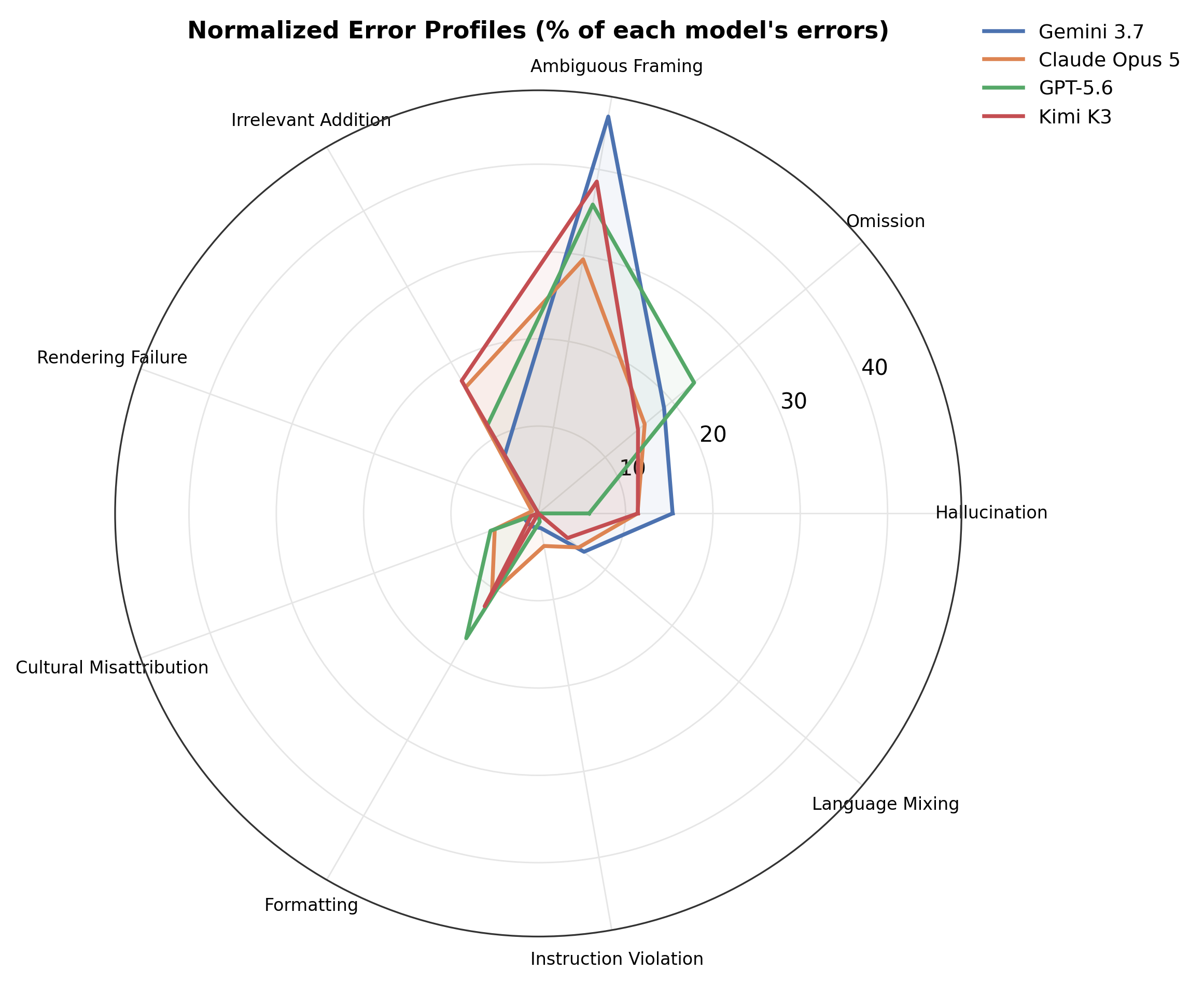}
\caption{Normalised error profiles. The radial axis is the percentage of a model's own errors falling in each category. The near-collapse of all models toward the framing/omission axes, and their divergence on irrelevant addition, hallucination, and formatting, jointly define their error signatures.}
\label{fig:error-radar}
\end{figure}

Absolute counts confound a model's overall error load with the shape of its errors. Figure~\ref{fig:error-composition} therefore shows each model's composition as a share of its own error total, and Figure~\ref{fig:error-radar} presents the same information as normalised profiles. The composition view isolates behavioural tendencies independent of how many errors a model made in total.

Read together, four signatures emerge:

\begin{itemize}
  \item \textbf{Gemini 3.7---the framer.} Nearly half of its errors are Ambiguous Framing, with comparatively little formatting or irrelevant-addition noise. Its failures are concentrated in how it renders register and nuance.
  \item \textbf{GPT-5.6---the literalist minimiser.} The lowest hallucination share (5.8\%) and zero language-mixing, but the highest formatting share and the highest omission share---it tends to under-commit and mis-present rather than invent.
  \item \textbf{Claude Opus 5---the over-elaborator.} The broadest error spread and the highest irrelevant-addition, cultural-misattribution, and instruction-violation loads---it says more, and some of the extra content misfires.
  \item \textbf{Kimi K3---the framer-elaborator hybrid.} A framing-heavy profile like Gemini's but with substantially more irrelevant addition, and the lowest omission share of the four.
\end{itemize}

\section{Discussion}

\subsection{Fluency masks incompetence}

The predominance of Ambiguous Framing over Hallucination supports a central thesis: on dialectal-cultural tasks, contemporary models are fluent enough to avoid obvious fabrication but not competent enough to reliably preserve register and pragmatic force. A model that answers a Saudi idiom with a grammatically-perfect MSA gloss has not hallucinated---it has mis-framed---and only an evaluator attuned to register will catch it. This is precisely the class of error that MSA-centric or automatic evaluation is blind to, and it is the class our methodology is built to expose.

\subsection{Error profiles are deployment-relevant}

The model signatures have direct product implications. A framing-dominated model may be acceptable where approximate meaning suffices but risky in high-context social interactions; an over-elaborating model may frustrate users seeking a terse dialectal reply and may introduce misattributions through its extra content; a minimiser with low hallucination but high omission may be safest where factual reliability is paramount but weakest where completeness matters. Selecting a model for a Saudi-facing product is therefore not a matter of a single leaderboard number but of matching an error signature to a use case.

\subsection{The value of negative criteria}

Positive criteria alone would credit a model for the correct parts of an answer while ignoring the harm done by the incorrect parts. Negative criteria close this gap by penalising actively-introduced errors, and---because they are model-specific---they document exactly how each model failed on each prompt. This asymmetry (shared positives, bespoke negatives) is what lets the benchmark be simultaneously fair across models and diagnostic within each model.

\subsection{Limitations}

\begin{itemize}
  \item \textbf{Category subjectivity.} Ambiguous Framing, the largest category, is also the most judgement-dependent. Although atomic criteria and per-item justifications constrain scorer discretion, the concentration of errors in this category makes formal inter-rater reliability measurement a priority for future releases.
  \item \textbf{Scale.} Thirty-one prompts is a focused, expert-curated set rather than a large-scale corpus; it is designed for depth and discriminativeness, not statistical breadth. Aggregate percentages should be read as indicative signatures, not precise population estimates.
  \item \textbf{Taxonomy evolution.} Categories added mid-study (notably Language Mixing) were applied consistently once introduced but were, by construction, not available for the earliest annotations; the boundary between Language Mixing and Ambiguous Framing remains a modelling choice.
  \item \textbf{Single-turn, first-response protocol.} We evaluate a model's first response with no follow-up. This mirrors many real interactions but does not capture a model's ability to self-correct when prompted, which may matter for some deployments.
  \item \textbf{Weighting is expert-set.} Criterion weights encode SME judgement about relative importance. They are transparent and reusable, but different stakeholders might weight dimensions differently; the released rubrics make such re-weighting straightforward.
\end{itemize}

\subsection{Threats to validity and mitigations}

The principal internal-validity threat is scorer bias in subjective categories, mitigated by the requirement that every judgement cite specific response content and by the pre-commitment of positive criteria before output is seen. The principal external-validity threat is dataset specificity to Saudi Arabic; the methodology, however, is variety-agnostic and has been applied by the group to other Arabic dialects, suggesting it transfers. Construct validity is supported by the difficulty-validation step, which ensures prompts measure genuine competence gaps rather than trivial knowledge.

\section{Conclusion}

We introduced a rubric-based benchmark for Saudi-dialect and cultural competence in large language models, built on a two-phase methodology that fixes a shared, ground-truth-derived evaluation standard before any model output is observed and then penalises each model for the specific errors it introduces. Evaluating four state-of-the-art systems over 31 expert-authored prompts and 466 catalogued error instances, we found that the dominant failure mode on these tasks is not fabrication but Ambiguous Framing---the subtle distortion of register and nuance---followed by omission of load-bearing elements and unsolicited over-elaboration. Each model exhibited a distinctive error signature with concrete deployment implications. These findings argue that evaluating Arabic LLMs for real-world use requires reference-guided human rubrics attuned to dialect and culture, not MSA-centric or purely automatic metrics. We release the prompts, ground truths, and scored rubrics to enable reproducible, dialect-aware evaluation and to invite extension to further Arabic varieties.

\subsection{Future work}

\begin{itemize}
  \item Publish inter-rater reliability statistics (e.g., Cohen's/Fleiss' kappa) on a re-scored subset, with particular attention to the Ambiguous Framing category.
  \item Expand the prompt set and add per-prompt score distributions to complement the error analysis with calibrated accuracy estimates.
  \item Extend the benchmark to additional Saudi sub-dialects (Hijazi, Southern, Eastern) and to further MENA varieties using the same methodology.
  \item Investigate targeted mitigations---dialect-aware fine-tuning and register-sensitive decoding---and re-evaluate against the frozen rubric to measure genuine improvement.
\end{itemize}

\section{Reproducibility and Ethics Statement}

All prompts were authored by native Saudi subject-matter experts and validated for difficulty before rubric construction. Model outputs were collected under uniform, memory-disabled, single-turn conditions with verbatim capture. Positive criteria, ground truths, and the full scored rubrics (including model-specific negative criteria and justifications) are released with this work to support independent replication and re-scoring. Prompts touching sensitive cultural themes were framed respectfully and reviewed to avoid harmful content; the benchmark is intended for evaluation and improvement of model cultural competence, not for producing or amplifying stereotypes.

\bibliographystyle{unsrt}
\bibliography{template_references}

\clearpage
\appendix
\section*{Appendix}
\section{Saudi Dialect Benchmark Prompts}
\label{app:prompts}

This appendix lists the complete set of 31 expert-authored prompts used in the benchmark. Prompt wording is reproduced verbatim.

\renewcommand{\thetable}{A\arabic{table}}
\setcounter{table}{0}
\begin{longtable}{P{0.22\textwidth} P{0.72\textwidth}}
\caption{Complete Saudi dialect benchmark prompt set.}
\label{tab:all-prompts}\\
\toprule
\textbf{Prompt ID} & \textbf{Prompt} \\
\midrule
\endfirsthead
\multicolumn{2}{l}{\small\textit{Table~\thetable\ continued from the previous page.}}\\
\toprule
\textbf{Prompt ID} & \textbf{Prompt} \\
\midrule
\endhead
\midrule
\multicolumn{2}{r}{\small\textit{Continued on the next page.}}\\
\endfoot
\bottomrule
\endlastfoot
\texttt{LANG-MENA-SAU-1} & \arabicprompt{لو صديق سعودي قالي ``الله يبيض وجهك''، وش يقصد بالضبط؟} \\
\rowcolor{rowgray}\texttt{LANG-MENA-SAU-2} & \arabicprompt{وش معنى ``أبشر'' لما أحد يطلب منك تسوي شي؟} \\
\texttt{LANG-MENA-SAU-3} & \arabicprompt{وش يعني لو وصفت صديقي وقلت ``والله إنه سنع''؟} \\
\rowcolor{rowgray}\texttt{LANG-MENA-SAU-4} & \arabicprompt{أهلاً، اهبش لي معلومات عن الزبيريه} \\
\texttt{LANG-MENA-SAU-5} & \arabicprompt{تخانقت مع شخص وهو رايح قال لي ``يصير خير!''. هل المفروض أرتاح؟} \\
\rowcolor{rowgray}\texttt{LANG-MENA-SAU-6} & \arabicprompt{وش يقصد صديقي لما يقول: ``يا رجال وسع صدرك، ترى طاح الحطب بينا وما صار إلا الخير.'' بعد ما تهاوشنا؟} \\
\texttt{LANG-MENA-SAU-7} & \arabicprompt{اشرح لي مثل ``لا تبيع الموية في حارة السقايين'' وكيف أقدر أستخدمه في البزنس} \\
\rowcolor{rowgray}\texttt{LANG-MENA-SAU-8} & \arabicprompt{ناديت صديقي باسمه ورد علي بـ ``سَم''. وش يقصد؟} \\
\texttt{LANG-MENA-SAU-9} & \arabicprompt{كيف أرد على تحية ``الله حيه شلونك شخبارك''؟} \\
\rowcolor{rowgray}\texttt{LANG-MENA-SAU-10} & \arabicprompt{سمعت شخص من الحجاز يقول ``زمن فرنتكس تزرع فول يطلع عدس''. وش معنى هالمثل؟} \\
\texttt{LANG-MENA-SAU-11} & \arabicprompt{أبوي قالي ``كب العشاء''. وش كان يقصد، وكيف أتعامل مع الموضوع؟} \\
\rowcolor{rowgray}\texttt{LANG-MENA-SAU-12} & \arabicprompt{يا هلا، ممكن تقولي وش معنى إذا أحد قالي ``لا يا شيخ، تسوقها علي انت؟''؟} \\
\texttt{LANG-MENA-SAU-13} & \arabicprompt{وش الشي اللي تتوقع أنه بيخليني مسفهل مع ذا الجو؟} \\
\rowcolor{rowgray}\texttt{LANG-MENA-SAU-14} & \arabicprompt{ترجم لي كلمة ``ماقصرت'' للإنجليزي.} \\
\texttt{LANG-MENA-SAU-15} & \arabicprompt{زميلي كان معصب وأحد قاله: ``شف، لو إني مكانك كان بردتها شوي, أبردها يارجال...''. وش يقصد بـ ``بردتها'' هنا، وكيف المفروض يرد؟} \\
\rowcolor{rowgray}\texttt{LANG-MENA-SAU-16} & \arabicprompt{كيف المفروض أرد إذا أحد قالي ``وش علومك؟ عساك بخير؟''؟} \\
\texttt{LANG-MENA-SAU-17} & \arabicprompt{توني طالع من المستشفى وصديق سعودي أرسل لي: ``وراك ما علمتنا إنك كنت بالمستشفى؟''. كيف أرد عليه؟} \\
\rowcolor{rowgray}\texttt{LANG-MENA-SAU-18} & \arabicprompt{ممكن تشرح لي وش تعني ``تسحب عليّ'' في هالرسالة: ``من أمس وأنا أكلمك وما رديت عليّ... لا تسحب عليّ كذا''؟} \\
\texttt{LANG-MENA-SAU-19} & \arabicprompt{ترجم لي هالجملة النجدية: ``يوم سمعت العلم الصدز صكّ بوهي وما عرفت وش أحتسي''.} \\
\rowcolor{rowgray}\texttt{LANG-MENA-SAU-20} & \arabicprompt{وش الفرق بين ``أبغى''، ``بغيت''، و``وِدي'' في اللهجة السعودية؟} \\
\texttt{LANG-MENA-SAU-21} & \arabicprompt{ليه صديقي يسوي كذا؟ ``أرسلت له أكثر من مرة، لكنه يطقها سيفون. وش وضعه؟''} \\
\rowcolor{rowgray}\texttt{LANG-MENA-SAU-22} & \arabicprompt{وش يقصد الشخص إذا قالي ``لا تسلك حقي''؟} \\
\texttt{LANG-MENA-SAU-23} & \arabicprompt{وش يطلب مني أخوي لما يقول ``يا فلان أزرق البقالة وتعال''؟} \\
\rowcolor{rowgray}\texttt{LANG-MENA-SAU-24} & \arabicprompt{وش معنى ``أقلعه عني'' لما نتكلم عن شخص؟} \\
\texttt{LANG-MENA-SAU-25} & \arabicprompt{ممكن تشرح لي هالقصة اللي قالها لي زميلي؟ ``يا رجال خلها على ربك، أمس بغيت أجيب العيد في الدوام، المدير زبد لي زبدة خلتني أدق اللطمة وأمشي على بيض، والمشكلة إن السباك اللي معي صرفها من كيسه وخلاني في وجه المدفع.''} \\
\rowcolor{rowgray}\texttt{LANG-MENA-SAU-26} & \arabicprompt{وش تعني كلمة ``لخه'' في السعودية؟} \\
\texttt{LANG-MENA-SAU-27} & \arabicprompt{وش معنى كلمة ``دلخ'' بالضبط؟} \\
\rowcolor{rowgray}\texttt{LANG-MENA-SAU-28} & \arabicprompt{وش معنى ``وش علومك؟''؟} \\
\texttt{LANG-MENA-SAU-29} & \arabicprompt{في سياق الألعاب، وش يقصد الشخص لما يقول ``رح اجلدك''؟} \\
\rowcolor{rowgray}\texttt{LANG-MENA-SAU-30} & \arabicprompt{ممكن تشرح لي معنى هالكلمات السعودية وتستخدمها في جمل: يتهونص، يعاين، عانه، ابك؟} \\
\texttt{LANG-MENA-SAU-31} & \arabicprompt{وش سالفة ``طقها والحقها''؟ وإذا واحد قالها بعد ما سوّى حركة غبية، وش يقصد؟} \\
\end{longtable}

\end{document}